\documentclass[letterpaper, 10 pt, conference]{ieeeconf}
\IEEEoverridecommandlockouts
\usepackage{cite}
\usepackage{amsmath,amssymb,amsfonts}
\usepackage{algorithmic}
\usepackage{graphicx}
\usepackage{textcomp}
\usepackage{xcolor}
\usepackage{multirow}
\usepackage{makecell}
\usepackage{booktabs}
\usepackage{tablefootnote}
\usepackage[normalem]{ulem}
\usepackage{comment}
\usepackage{graphicx}

\def\BibTeX{{\rm B\kern-.05em{\sc i\kern-.025em b}\kern-.08em
    T\kern-.1667em\lower.7ex\hbox{E}\kern-.125emX}}

\title{
Personalized Stress Monitoring using Wearable Sensors in Everyday Settings
}

\author{
Ali Tazarv$^{\|}$, 
Sina Labbaf$^{\ast}$, 
Stephanie M. Reich$^{\dagger}$, 
Nikil Dutt$^{\ast,\mathsection}$, 
Amir M. Rahmani$^{\ddagger,\ast,\mathsection}$, 
Marco Levorato$^{\ast}$\\
${\|}$ Dept. of Electrical Engineering and Computer Science,
${\ast}$ Dept. of Computer Science,\\
${\dagger}$ School of Education,
${\ddagger}$ School of Nursing,
${\mathsection}$ Institute for Future Health (IFH)\\
University of California, Irvine
}

\newcommand\blfootnote[1]{%
  \begingroup
  \renewcommand\thefootnote{}\footnote{#1}%
  \addtocounter{footnote}{-1}%
  \endgroup
}

\begin{document}

\maketitle

\thispagestyle{empty}
\pagestyle{empty}


\begin{abstract}
Since stress  contributes to a broad range of mental and physical health problems,
the objective assessment of stress is essential for behavioral and physiological studies.
Although several  studies have evaluated  stress levels in controlled settings, objective stress assessment in everyday settings is still largely under-explored due to challenges arising from confounding contextual factors and limited adherence for self-reports.
In this paper, we explore the objective prediction of stress levels in everyday settings based on heart rate (HR) and heart rate variability (HRV) captured via  low-cost and easy-to-wear photoplethysmography (PPG) sensors that are widely
available on newer smart wearable devices.
We present a layered system architecture for personalized stress monitoring that supports a tunable collection of data samples for labeling, and present a method for selecting informative samples from the stream of real-time data for labeling. 
We captured the stress levels of fourteen volunteers through self-reported questionnaires over periods of between 1-3 months, and explored binary stress detection based on HR and HRV using Machine Learning Methods.
We observe promising  preliminary results given that the dataset is collected in the challenging environments of everyday settings. 
The binary stress detector is fairly accurate and can detect stressful vs non-stressful samples with a macro-F1 score of up to \%76.
Our study lays the groundwork for more sophisticated labeling strategies that generate context-aware, personalized models that will empower health professionals to provide personalized interventions.

\end{abstract}


\section{Introduction}
\label{sec: intro}
\blfootnote{This work was partially supported by NSF Smart and Connected Communities (S\&CC) grant CNS-1831918.}

Stress can contribute to illness through its direct physiological effects or indirectly through maladaptive health behaviors (\emph{e.g.}, smoking, poor eating or sleeping habits)~\cite{glanz2008stress}. It is therefore critical to motivate people to adjust their behavior and lifestyle and introduce appropriate strategies to achieve a better stress balance before an increased level of stress results in serious health conditions \cite{bakker2011s}.

The increasing availability of wearables, interconnected devices capable of acquiring high-quality biosignals, opens important opportunities for advanced machine learning-enabled health monitoring and intervention applications \cite{FIROUZI2018583,MIERONKOSKI201778}. 
Recent literature~\cite{han2020objective} demonstrates that it is indeed possible to objectively detect stress by analyzing biological signals. 
However, existing objective stress detection frameworks are designed for controlled settings, where data is recorded while users are in a set of predefined physical states or performing certain activities (\emph{e.g.} sitting, lying down, running). 
On the other hand, stress detection needs to be performed in \textit{everyday settings}, where subjects are engaged in their normal daily activities and routines.
Everyday settings pose  inherent challenges for stress monitoring, including: 
real-time collection and analysis of data;  the lower quality of signals due to motion and noise artifacts (MNAs); and difficulties in collecting self-reports due to limited user adherence~\cite{han2020objective,NAEINI2019551}.
Furthermore, personalization of stress monitoring in
everyday settings raises additional challenges:
features may emerge that are specific to the user's characteristics, behavioral patterns, physiology and context, as well as sensor setup/configuration, thus presenting a much higher degree of variations from one person to another compared to controlled settings. These differences can result in degrading the performance of general classifiers in everyday settings.

Since effective everyday stress monitoring and intervention 
must be personalized and context-aware,
the underlying machine learning core needs to be adapted to
match the stream of data generated by the user. 
It is important to note that personalized classifiers often outperform those trained using data from the general population even in idealized settings \cite{han2020objective}.
%
These classifiers depend on labels generated from subjects in-the-moment to accurately record instances of stress.
However, users may not respond in a timely or interactive manner, resulting in a trade-off between the number of labels provided by the subjects versus the accuracy of the predictive model.
This trade-off creates the need for a smart label query strategy that we use to  explore real-time stress detection based on wearable data in every day settings.

The key contributions of this paper are:

\noindent
$\bullet$ We architect a three-tier system for the collection and real-time analysis of biosignals labeled using self-reports. The system is composed of wearable sensors, a smartphone serving as a gateway, and a cloud server.
We  discuss system-level challenges influencing data acquisition capabilities.

\noindent
$\bullet$ We  develop a smart strategy to obtain labels for an adequate number of samples to proportionally represent the entire data from each user while capturing less overlapping regions of the feature space.
 
\noindent
$\bullet$ We develop a  machine learning based stress predictor. We map the stress labels into binary \textit{stressed (1)} and \textit{not stressed (0)}, and then train classification methods using these labels. 

\noindent
$\bullet$ 
We capture the stress levels of fourteen volunteers through self-reported questionnaires over periods of between 1-3 months, and evaluate our binary stress detection based on HR and HRV.
Our classifiers are able to identify the binary classes with an F1 score up to 76\%. We also analyze the effect of personalization, and show how the stress detection performance improves over time, as we collect more labels from a subject and use those in the training process.

The rest of the paper is organized as follows.
Section \ref{sec: system_model}  describes  the system architecture we used for data collection, and the proposed strategy for label collection in everyday settings. 
Section \ref{sec: stress_detection} describes the classification methods we used for detecting stress. 
Section \ref{sec: results} presents our analysis and results on stress detection.
Section \ref{sec: conclusion} concludes the paper
with a summary and directions for future work.

\section{System Model and Data Collection}
\label{sec: system_model}

The ultimate goal of data collection is to train a personalized classifier for stress detection based on biosignals (PPG).
One of the key challenges in collecting such datasets in everyday settings is the interaction with the users, as sending queries for labeling too often may overwhelm the users and may also lead to unnecessary data collection.

To enable real-time interaction with the user while minimizing resource usage, a layered design is necessary.
The sensor layer should collect the raw signals, while the cloud layer processes the data, determines the quality of the signals, performs feature extraction and other computationally intensive and power consuming tasks. We also have to provide users with an interface to input their labels and build a communication path between these layers. 
In this section, we present our proposed solution for these requirements.

The experimental procedures involving human subjects described in this paper were approved by the Institutional Review Board (IRB) at UC Irvine. 
\subsection{System Architecture}
\begin{figure}[tbp]
    \centering
    \includegraphics[width=1\linewidth]{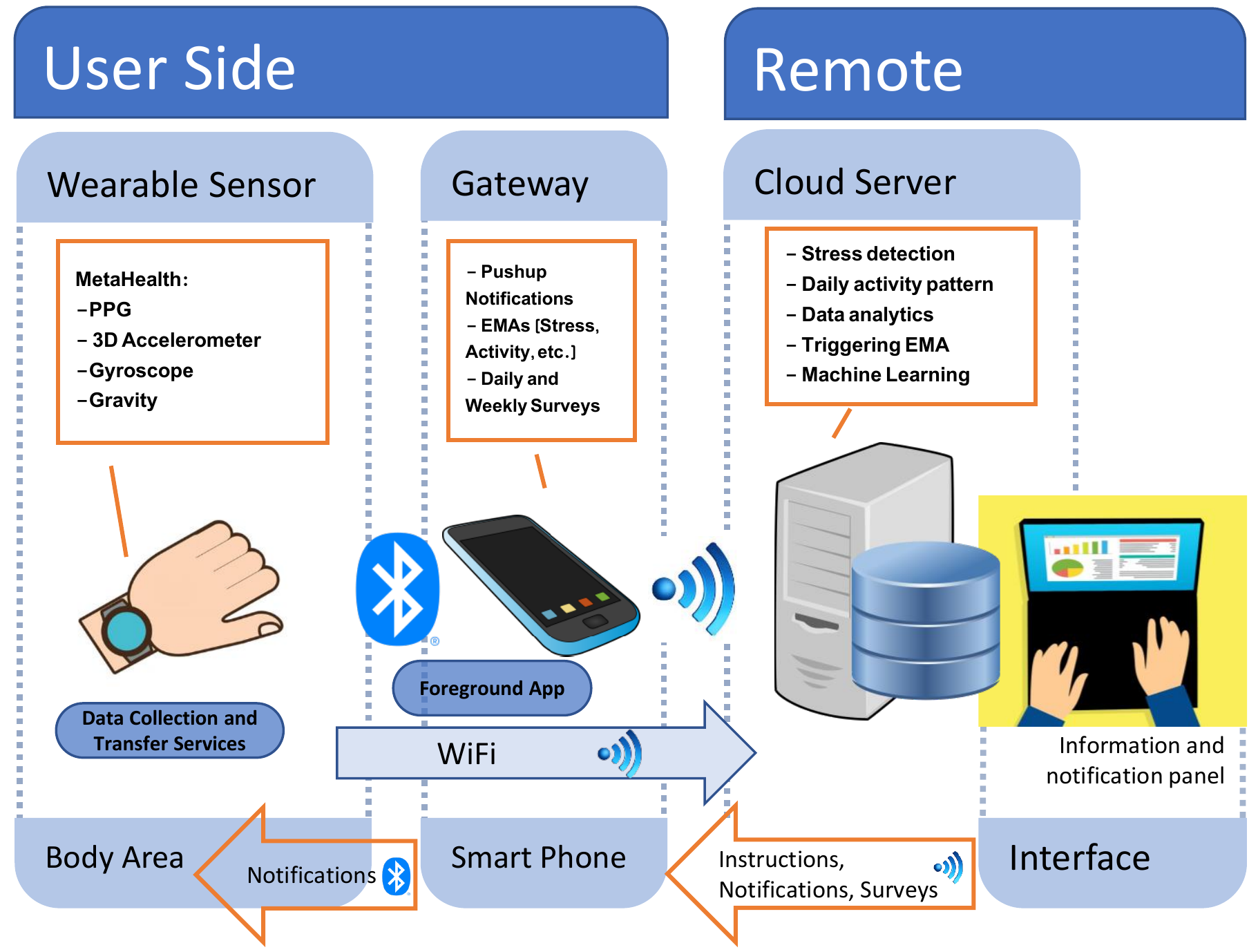}
    \caption{Overview of the system architecture.}
    \label{fig: architecture}
    \vspace{-3ex}
\end{figure}

Figure~\ref{fig: architecture} outlines the system we developed to acquire the real-time dataset associating self-reported stress ratings from users to biosignals from wearable devices.
We use a 3-tier architecture composed of a wearable device (sensor layer), a smartphone (edge layer), and a remote cloud server (cloud layer) working to collect and process the data, as described below.

\subsubsection{\textbf{Sensor Layer}}
The wearable platforms acquire and transmit raw physiological (PPG) and Movements (Accelerometer, Gyroscope and Gravity) signals. 
We used Samsung Gear Sport smartwatches and developed a service running on Tizen operating system to collect raw PPG and movement signals. The sampling frequency of all the above mentioned signals is $20 Hz$. The watch can send the data directly to the cloud layer (if connected to a local Wi-Fi), or to the smartphone via Bluetooth.
The raw signal acquisition application includes two services and a user interface (UI). The first service collects the sensor data at a constant rate (once every 15 minutes) and duration (2-minutes intervals) and sends it to the cloud. If the service fails to send the data to the cloud immediately, the data will be stored on the watch and transferred to the server at a later time. The UI is a simple app installed on the watch for restarting these two services. 
\subsubsection{\textbf{Cloud Layer}}
A cloud web-server receives the data samples from the watch and processes them.
Based on the observed features of each incoming sample, an internal logic (described later) determines whether it needs to ask users for a label or not. The responses from users are transferred to the cloud and stored in the database.

\subsubsection{\textbf{Edge Layer}}
We developed a smartphone app that asks the participants for labels through an Ecological Momentary Assessment (EMA). The EMA is triggered by the cloud for a portion of samples and when triggered, a push notification is displayed on the phone that asks the participant about their stress levels and recent physical activity or state (e.g., sitting, standing, etc.). The stress levels in the EMA are \textit{not at all}, \textit{a little bit}, \textit{some}, \textit{a lot}, and \textit{extremely}.

The edge-cloud connection is established through the Internet on the smartphone. In addition, the watch is connected to the smartphone using Bluetooth Low Energy (BLE). In order to send the collected data to the cloud, the watch proxies the connected phone's Internet connection through BLE. This setting is energy efficient, and thus suitable for everyday setting applications. This is a back up connection and takes effect when the watch is not directly connected to the Wi-Fi.

\subsection{Data Labeling}
\label{sec: data_labeling}
The system needs to parsimoniously trigger the EMA to collect labels to build a meaningful dataset as quickly as possible without imposing excessive burden on the user. Therefore, we devised a selection method to select informative samples to be labeled by the user.
However, before applying the selection method, the raw signals need to be pre-processed for extracting the corresponding features.
 
\subsubsection{Data Cleaning and Feature Extraction}
When a raw PPG sample is received at the cloud, we first filter the raw signal to clean up the high and low frequency noises. We apply a Butterworth band-pass filter of order 3, with cut off frequencies set at $(0.7Hz$, $3.5Hz)$, corresponding to $42bpm$ and  $210bpm)$ respectively. Then, we pass the signal through a moving average filter and at the end apply a peak detector on it. Using the peak points of the filtered signal,
we extract thirteen features from each sample (2 minutes of data). These features are:
BPM, IBI, SDNN, SDSD, RMSSD, pNN20, pNN50, MAD, SD1, SD2, S, SD1/SD2, and BR\footnote{
\textbf{BPM:} Beats per Minute, Heart Rate.
\textbf{IBI:} Inter-Beat Interval, average time interval between two successive heart beats (called NN intervals).
\textbf{SDNN:} Standard Deviation of NN intervals.
\textbf{SDSD:} Standard Deviation of Successive Differences between adjacent NNs.
\textbf{RMSD:} Root Mean Square of Successive Differences between the adjacent NNs.
\textbf{pNN20:} The proportion of successive NNs greater than 20ms (or 50ms for pNN50).
\textbf{MAD:} Median Absolute Deviation of NN intervals.
\textbf{SD1 and SD2:} Standard Deviations of the corresponding Poincaré plot.
\textbf{S:} Area of ellipse described by SD1 and SD2.
\textbf{BR:} Breathing Rate.}.
We use these features for further processing and decision making.

\subsubsection{Strategy for Labelling Selected Data}
Data collection consists of two phases:

\noindent
\textbf{Initial Phase:} 
In order to get an initial estimate of the distribution of samples in the sample space, we start the procedure by observation. For the first \textit{N} samples (100 samples in our setup; $\sim$25 hours of wearing the watch), we do not trigger any EMAs. At the end of this phase, we get an estimate of the distribution of samples in the samples space.

\noindent
\textbf{Query Phase:}
For samples after the initial phase (\textit{N+1} and above), we trigger the EMA (ask for labels) for a portion of samples.
The probability of selecting each sample to be labeled is proportional to the number of previous samples (unlabeled) in its neighborhood. This way, if a sample falls in a region in which there has been a large number of unlabeled samples, 
it is more likely that we ask the user for the label. For each region after we collect sufficient number of labeled samples, we stop collecting labels. However, the minimum probability of triggering the EMA for a sample is \textit{P} = 0.1. This means if a sample falls in a region where there is little or no previous samples, the probability of query is still non-zero. This results in exploring unseen regions, as well as regions with higher densities.

We  capture  the  stress  levels  of  fourteen  volunteers  through
self-reported   questionnaires   over   periods   of   between   1-3 months.
The total number of samples, along with the number of labeled samples for each user is presented in Table \ref{tab: data_count}.
\begin{table}[bt]
    \centering
    \vspace{.15cm}
    \caption{Number of samples and labels collected from each subject}
    \begin{tabular}{l r r r} 
    \hline \\[-1.5ex]
    Subject & \# samples & total labels & used labels\\
    \hline\\
    S01  & 4,580 & 228 & 217 \\
    S02  & 2,164 & 101 & 92  \\
    S03  & 1,764 & 67  & 42  \\
    S04  & 2,580 & 56  & 53  \\
    S05  & 2,267 & 68  & 59  \\
    S06  & 17,552& 376 & 370 \\
    S07  & 10,087& 105 & 101 \\
    S08  & 2,752 & 96  & 93 \\
    S09  & 1,236 & 53  & 50 \\
    S10 & 7,910 & 119 & 104 \\
    S11 & 2,555 & 73  & 60  \\
    S12 & 1,2296& 956 & 942 \\
    S13 & 3,738 & 47  & 45  \\
    S14 & 1,332 & 61  & 55  \\
    \vspace{-5ex}
    \end{tabular}
  \label{tab: data_count}
\end{table}
\section{Stress Detection}
\label{sec: stress_detection}
In this section, we explore the possibility of predicting stressful vs non-stressful moments based on the collected signals. We train our stress detection models on both personal and general datasets to evaluate the performance. For stress detection, we use several machine learning classification algorithms such as Multi Layer Perceptron (MLP), Random Forest (RF), k-Nearest Neighbors (kNN), Support Vector Machine (SVM), and XGBoost.
MLP is a class of feed-forward neural networks that can be trained to do nonlinear classification and regression tasks. RF is an ensemble learning method for classification that operates by constructing a number of decision trees at training time and outputting the class that is the mode of the classes of the individual trees. kNN uses the \textit{k} nearest points and takes a majority vote to predict the class of the sample. SVM finds hyper planes and partitions the sample space into different classes. XGBoost is an implementation of Gradient Boosted Decision Trees that is fast and performs well in classification tasks. 
We train each of these classifiers on our dataset and analyse their performance using machine learning methods and F1 score as the evaluation metric.

\section{Results and analysis}
\label{sec: results}
Detecting stress by only using PPG signals in everyday settings is a challenging task \cite{han2020objective}. To evaluate the validity of our models, we perform K-Fold cross validation (\textit{K} = 5). In K-Fold Cross Validation, we split the data into K equally sized segments (folds) and in each iteration use 1 fold for evaluation, and the other K-1 folds for training. The data is stratified prior to be split in K folds, to ensure each fold is a proper representative of the whole. The ML methods we used are introduced and explained in Section \ref{sec: stress_detection}. To evaluate the performance of each method on our collected dataset, we used Macro-F1 score\footnote{F1 score is defined on each class separately. macro-F1 score is the average of F1 scores on all the classes (two here).}.
The mean and standard deviation ($\mu \pm \sigma$) of the F1 scores over the K folds are presented in Table \ref{tab: classification}. Based on these experiments, RF outperforms other methods for most cases (except for the first row).

\begin{table*}[bt]
    \centering
    \vspace{.15cm}
    \caption{Mean value of F1 score for 5 fold cross validation ($\pm$ standard deviation) for stress detection \\ based on PPG features, baseline is always "not at all" class}
    \scalebox{1}{
    \begin{tabular}{r l l l l l l } 
    \specialrule{0.1em}{0.1em}{.1em} \\[-1.5ex]
    &   & \multicolumn{5}{c}{5 fold Cross Validation, F1 Score}\\
    \cline{3-7}\\[-1.5ex]
    Binary Classes & Number of Samples & MLP & SVM & kNN & RF & XGBoost\\[1ex]
    \hline \\[-1ex]
    \textit{a little bit} (1) VS. baseline (0):         & (605, 143) & \textbf{0.73 ± 0.06} & 0.66 ± 0.03 & 0.72 ± 0.03 & 0.67 ± 0.04 & 0.72 ± 0.04\\
    \textit{some} (1) VS. baseline (0):                 & (299, 143) & 0.70 ± 0.03 & 0.69 ± 0.04 & 0.66 ± 0.06 & \textbf{0.71 ± 0.05} & 0.70 ± 0.04\\
    \textit{a lot} or \textit{extremely} (1) VS. baseline (0):  & (72, 143)  & 0.68 ± 0.06 & 0.69 ± 0.13 & 0.69 ± 0.04 & \textbf{0.76 ± 0.05} & 0.73 ± 0.09\\
    \textit{some, a lot} or \textit{extremely} (1) VS.  &&&&&&\\
    \textit{a little bit} or \textit{not at all} (0):   & (748, 371) & 0.62 ± 0.04 & 0.60 ± 0.04 & 0.59 ± 0.03 & \textbf{0.63 ± 0.02} & 0.63 ± 0.04\\[1ex]
    \hline\\
    \vspace{-6ex}
    \end{tabular}}
   \label{tab: classification}
\end{table*}


\subsection{Personalization}
The bias in the physiological data (both the signals and the labels) can be different for personal or general datasets \cite{han2020objective}. Therefore, we show the effect of personalization and how it improves the prediction accuracy on our collected dataset. To this end, we consider 3 participants from which more data is available (subjects S06, S10, S12). In the first step, we exclude the data from one subject (e.g. S06), train on the data from all other subjects, then test on half of the data from S06 (picked randomly). In the next step, in order to personalize the model, we use the other half of data from S06 and use it for training (along with the data from other users), and then test it on the first half of the data from S06. The results are reported in Table \ref{tab: personalization}. As can been seen from these results, personalization improves the prediction performance (macro-F1 score).
\begin{table}[bt]
    \centering
    \vspace{.15cm}
    \caption{Effect of personalization on stress prediction performance. Before and after rows are F1 scores before and after personalization for each user.}
    \scalebox{1}{
    \begin{tabular}{r l l l l l l} 
    \specialrule{0.1em}{0.1em}{.1em} \\[-1.5ex]
    && \multicolumn{5}{c}{F1 Score on 50\% of data from one user}\\
    \cline{3-7}\\[-1.5ex]
    \textbf{User} & Personalization & MLP & SVM & KNN & RF & XGB\\[1ex]
    \hline \\[-1ex]
    \multirow{2}{*}{S06:}  
    & before & 0.43& 0.37 & 0.44 & 0.40 & 0.38\\
    & after & \textbf{0.53} & \textbf{0.54} & \textbf{0.50} & \textbf{0.54} & \textbf{0.52}\\
    \hline\\[-1ex]
    \multirow{2}{*}{S10:}   
    & before & 0.58 & 0.63 & 0.60 & 0.612& 0.60\\
    & after & \textbf{0.62} & 0.62 & \textbf{0.63} & \textbf{0.616}& \textbf{0.61}\\
    \hline\\[-1ex]
    \multirow{2}{*}{S12:}   
    & before & 0.58& 0.59 & 0.61 & 0.59 & 0.55\\
    & after & \textbf{0.63} & \textbf{0.62} & \textbf{0.64} & \textbf{0.63} & \textbf{0.61}\\[1ex]
    \hline\\
    \vspace{-5ex}
    \end{tabular}}
   \label{tab: personalization}
\end{table}

\subsection{Improvement Over Size of Training Data}
\begin{figure}[tbp]
    \centering
    \includegraphics[width = 1.1\linewidth]{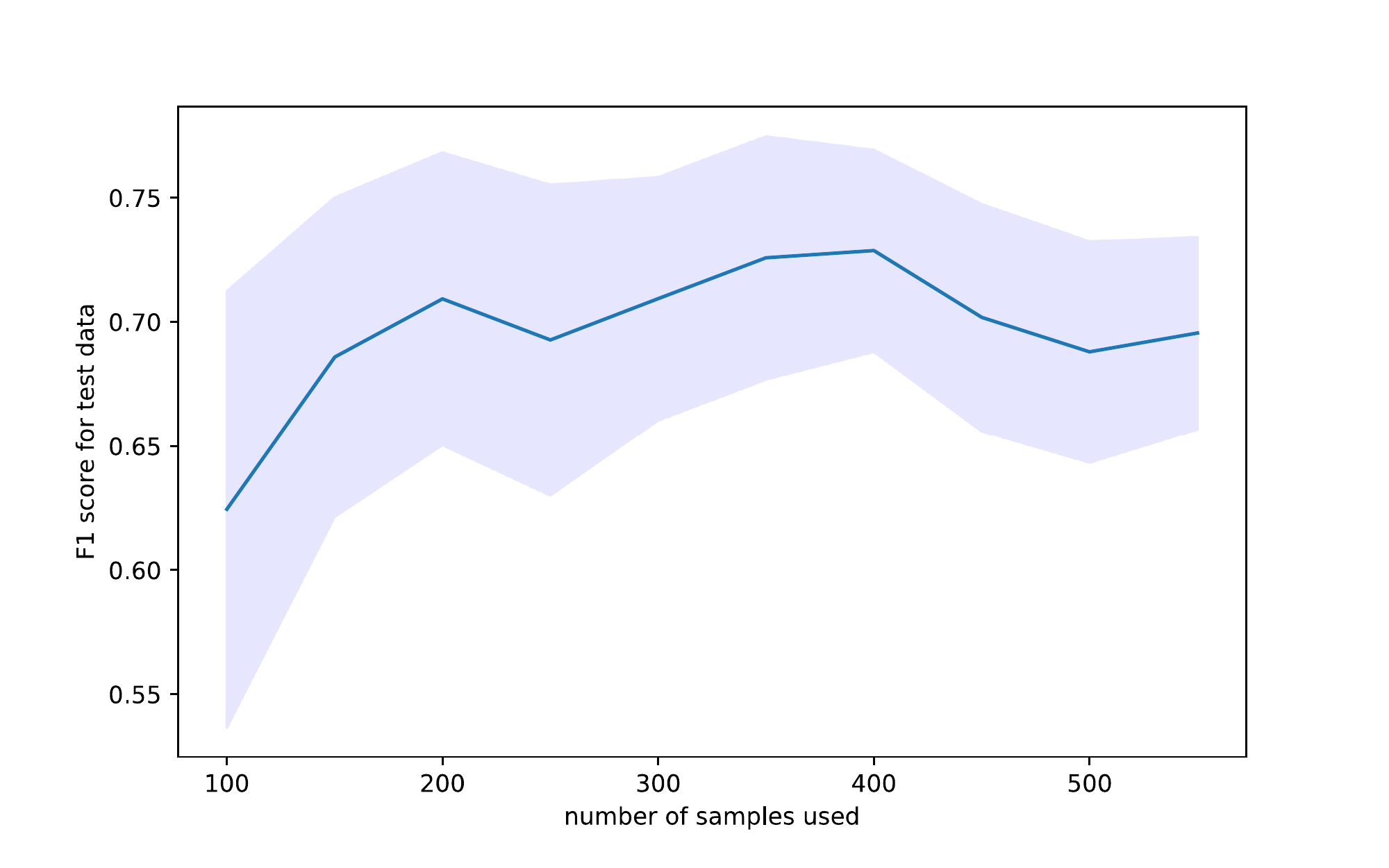}
    \vspace{-3ex}
    \caption{Stress Prediction over size of training data}
    \vspace{-2ex}
    \label{fig: prediction_progress}
\end{figure}

As we collect more labels from a user, the stress prediction can be performed more accurately. In order to show this process, we only use data from one user having for him a large number of labels is available (Subject S12). We also use random forest classifier for this experiment. We randomly separate 100 samples and use those for testing. We use the rest of the data in an incremental manner; first we train the model on 100 samples only, and then increase the training size. In each step, we test the trained model on the test data (all from S12). The improvement of prediction performance is presented in Figure \ref{fig: prediction_progress}. The model improves as we increase the subject's data size up to about 300 samples. For each step, we repeat the process (selection of test data and training data) 100 times, and the values in the figure show the mean and the standard deviation of the F1 score over all these 100 experiments. 

\section{Conclusions and Future Work}
\label{sec: conclusion}
Collecting photoplethysmogram (PPG) signals with enough labels collected from users in everyday settings is a challenging task. 
Our study used a Samsung Gear Sport smartwatch as a wearable device for data collection and utilized a method to improve the labeling procedure.
The data were collected from fourteen active volunteers.
We explored the possibility of detecting stressful vs non stressful moments (samples) using leave-samples-out validation based on PPG signals, in everyday settings. We analyzed the improvement of the trained classifier, as we personalize the classifier with samples from a certain subject.
The results are promising: we achieved macro-F1 scores up to 76\% for binary classification of stressful vs non stressful samples.
This motivates our future work to utilize  more advanced methods, possibly variants of active learning, in the labeling procedure. 
More informative labels will allow us to design classifiers that can possibly detect mental health conditions of users based on HRV from their biosignals and the type of user activity -- promising to provide valuable tools for mental health professionals to better diagnose and treat stress and anxiety in a personalized way.

\bibliographystyle{ieeetr}
\bibliography{main}
\vspace{12pt}
\end{document}